\documentclass[runningheads]{llncs}
\usepackage{graphicx}
\usepackage{cite}
\usepackage{algorithmic}
\usepackage{graphicx, caption}
\usepackage{textcomp}
\usepackage{makecell}
\usepackage{comment}
\usepackage{etoolbox}

\usepackage{subfigure}
\usepackage{tabularx}

\usepackage{color}
\definecolor{mygreen}{RGB}{84, 130,53}
\definecolor{myred}{RGB}{192, 0, 0}
\definecolor{myblue}{RGB}{0, 112, 192}
\definecolor{myyellow}{RGB}{255, 192, 0}

\begin{document}
\title{An empirical evaluation of AMR parsing for legal documents}
%
%
\author{Vu Trong Sinh \and
Nguyen Le Minh
}
%
%
\institute{Japan Advanced Institute of Science and Technology (JAIST) \\
\email{sinhvtr@jaist.ac.jp}, \email{nguyenml@jaist.ac.jp}}
%
\maketitle              
\begin{abstract}
Many approaches have been proposed to tackle the problem of Abstract Meaning Representation (AMR) parsing, helps solving various natural language processing issues recently. In our paper, we provide an overview of different methods in AMR parsing and their performances when analyzing legal documents. We conduct experiments of different AMR parsers on our annotated dataset extracted from the English version of Japanese Civil Code. Our results show the limitations as well as open a room for improvements of current parsing techniques when applying in this complicated domain.

\end{abstract}
\section{Introduction}\label{intro}
In Natural Language Processing, semantic representation of text plays an important role and receives growing attention in the past few years. Many semantic schemes have been proposed, such as Groningen Meaning Bank \cite{BasileBosEvangVenhuizen2012LREC}, Abstract Meaning Representation \cite{Banarescu2013}, Universal Conceptual Cognitive Annotation \cite{Abend_universalconceptual}. In which, Abstract Meaning Representation (AMR) has shown a great potential and gained popularity in computational linguistics \cite{Wang2017} \cite{Peng2017AddressingTD} \cite{Flanigan2016} \cite{Kontas2017}. 

AMR is a semantic representation language that encodes the meaning of a sentence as a rooted, directed, edge-labeled, leaf-labeled graph while abstracting away the surface forms in a sentence. Every vertex and edge of the graph are labeled according to the sense of the words in a sentence. AMR can be represented in PENMAN notation, for a human to read and write easily, or graph structure, for a computer to store in its memory, or decomposed into conjunctions of logical triples, for calculating the difference among AMRs. Table \ref{tab:format} shows an example of AMR annotation for the sentence \textit{"The boy wants to go"} with different formats mentioned above.

\begin{table}
    \caption{Abstract meaning representation for the sentence \textit{"The boy wants to go"} in three formats}
    \label{tab:format}
    \begin{tabular}{|m{4.7cm}|m{3cm}|m{4.1cm}|}
        \hline
        $\exists w, b, g: instance(w, want-01) \newline
        \wedge instance(g, go-01) \wedge instance(b, boy) \newline 
        \wedge arg0(w, b) \wedge arg1(w, g) \wedge arg0(g, b) $ &
        (w / want-01 \newline \hspace*{0.5cm} :arg0 (b /boy) \newline \hspace*{0.5cm} :arg1(g / go-01) \newline \hspace*{1cm} :arg0 b) & \raisebox{-\totalheight}{\includegraphics[width=0.33\textwidth, trim= -1 -5 0 0]{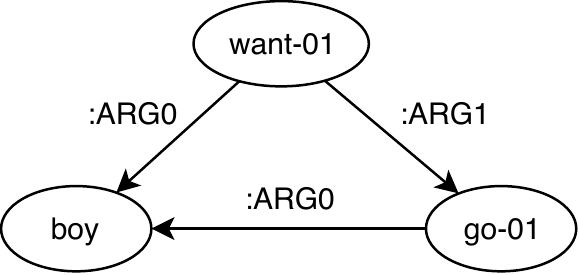}}\\
        \hline
    \end{tabular}
\end{table}{}

AMR has been applied as an intermediate meaning representation for solving various problems in NLP including machine translation \cite{Jones2012}, summarization \cite{liu2015toward}, event extraction \cite{Pan2015UnsupervisedEL}, \cite{Rao2017BiomedicalEE}, \cite{Huang2016LiberalEE}, machine comprehension \cite{Sachan2016}. For AMR to be useful in these problems, the AMR parsing task, which aims to map a natural language string to an AMR graph, plays a crucial role. Despite the advantages in handling semantic attributes of text, there are not many works exploring the application of AMR in analyzing legal documents. Unlike other domains, understanding legal text faces a number of challenges due to the special characteristics such as complicated structures, long sentences, domain-specific terminology. 

In this paper, we would like to investigate the potential of AMR in this interesting field. We provide an overview of main approaches in current AMR parsing techniques in section \ref{approaches}. From each approach, we choose best systems that already published the source code to conduct our experiments. In section \ref{dataset}, we revise the dataset \textbf{JCivilCode-1.0} introduced in 2017 by Lai et al. \cite{Lai2017} with some modifications and additional samples. We also extract sentences with various lengths from a well-known dataset LDC2017T10 \footnote{https://catalog.ldc.upenn.edu/ldc2017t10} in common domain to have more observation on the performances of each system. Our results and some discussions are provided in section \ref{experiments}.

\section{Approaches in AMR parsing}\label{approaches}

\subsection{AMR notation}

In AMRs, each node is named by an ID (variable). It contains the semantic concept, which can be a word (e.g. \textit{boy}) or a PropBank frameset (e.g. \textit{want-01}) or a special keyword. The keywords consist of entity type (e.g. \textit{date-entity, ordinal-entity, percentage-entity}), quantities (e.g. \textit{distance-quantity}), and logical conjunction(e.g. \textit{and, or}). The edge between two vertices is labeled using more than 100 relations including frameset argument index (e.g. \textit{“:ARG0”, “:ARG1”}), semantic relations (e.g. \textit{“:location”, “:name”}), relations for quantities (e.g. \textit{:quant, :unit, :scale}), relations for date-entities, relations for listing (e.g. \textit{:op1, :op2, :op3}). AMR also provides the inverse form of all relations by concatenating -of to the original relation (e.g. :location vs :location-of ). Hence, if \textit{r} is a directed relation of two entities \textit{a} and \textit{b}, we have $R(a, b) \equiv R-of(b, a)$. This inverse relation helps keep the focus on the entity instead of the verb sense as default. 

\begin{figure*}
  \centering
    \subfigure[Alignment-based method]
    {
        \includegraphics[width=0.35\textwidth]{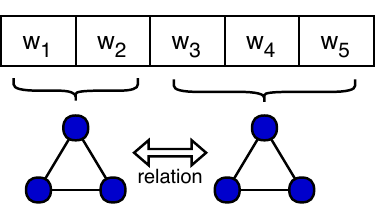} 
        \label{fig:alignment}
    }
    \subfigure[Grammar-based method]
    {
        \includegraphics[width=0.35\textwidth]{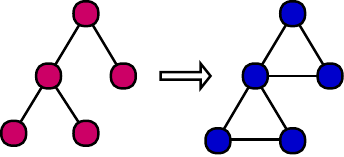}
        \label{fig:grammar}
    }
    \newline
    \subfigure[Machine-translation-based method]
    {
        \includegraphics[width=0.5\textwidth]{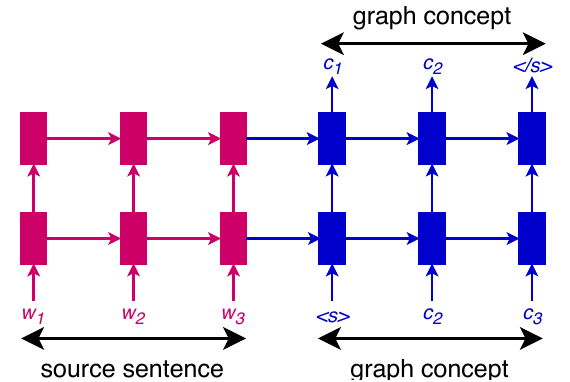}
        \label{fig:mt}
    }
  \caption{Main approaches in AMR parsing}
  \label{fig:exph,exgr}
\end{figure*}

The task of parsing a natural language text into an AMR graph faces a lot of challenges, such as word-sense disambiguation, semantic graph construction, data sparsity. Many approaches have been proposed to tackle this problem. They can be divided into three main categories: alignment-based, grammar-based and machine-translation-based.

\subsection{Alignment-based parsing}

One of the pioneer AMR parsing solutions is \textbf{JAMR} introduced by Flanigan et al. in 2014\cite{Flanigan_adiscriminative}, which build a two-part algorithm that first identifies concepts with an automatic aligner and then identifies the relations that it obtains between these by searching for the maximum spanning connected subgraph from an edge-labeled, directed graph representing all possible relations between the identified concepts. This method provided a strong baseline in AMR parsing. Follow this approach, Zhou et al. \cite{Zhou2016} extended the relation identification tasks with a component-wise beam search algorithm. Chunchuan and Titov \cite{Chunchuan_LatentAlignment} improved this method by considering alignments as latent variables in a joint probabilistic model. They used variational autoencoding technique to perform the alignment inference and archieved the state-of-the-art in AMR parsing until now. But the source code for this model has not been published completely yet. In this paper, we take the JAMR model \cite{Flanigan_adiscriminative} to analyze and conduct experiments.

\begin{figure}
\centerline{\includegraphics[width=2in]{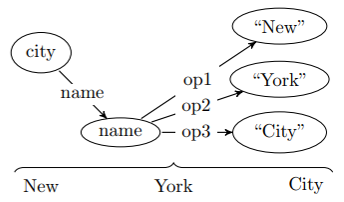}}
\caption{Alignment of the words span "New York City" with AMR fragment \cite{Flanigan_adiscriminative}}
\label{fig:jamr}
\end{figure}

The core idea of alignment-based methods is to construct a concept set by aligning the Propbank concepts with the words that evoke them. The authors build an automatic aligner that uses a set of rules to greedily align concepts to words. The authors use WordNet to generate candidate lemmas and a fuzzy match of a concept, defined to be a word in the sentence that has the longest string prefix match with that concept's label. For instance, the fuzzy match for \textbf{apply-01} could be aligned with \textit{"application"} if this word is the best match in the sentence. Figure \ref{fig:jamr} shows an example of aligning words in a sentence with AMR concepts. This JAMR aligner is widely used in many later works.

In the first stage of identifying concepts, given a sentence $w=\{w_1, w_2, ..., w_n \}$, the parser segments w into subsequences, denoted $\{w_{b_0:b_1}, w_{b_1:b_2}, ..., w_{b_{k-1}:b_k}\}$, called contiguous spans. A span $\{w_{b_{i-1}:b_i}\}$ is then assigned to a concept graph fragment $c_i$ from the concept set $clex\{w_{b_{i-1}:b_i}\}$, or to $\theta$ for words that evoke no concept. This assigning between sequence of spans $b$ and concept graph fragment $c$ is calculated by a score function:

\begin{equation}
score(b, c; \theta) = \sum_{i=1}^{k}\theta^{T}f(w_{b_{i-1}:b_i}, b_{i-1}, c_i),
\end{equation}
where $f$ is a feature vector representation of a span and one of its concept graph fragments in the sentence. The features can be fragment given words, length of the matching span, name entity recognizing or bias.

To find the highest-scoring between $b$ and $c$, JAMR uses a semi-Markov model. Let $S(i)$ be the score of the first $i$ words of the sentence ($w_{o:i}$). Then $S(i)$ is calculated recurrently via the previous scores, with the initialization $S(0)=0$. Obviously, $S(n)$ becomes the best score. To obtain the best scoring concept labeling, JAMR uses back-pointers method, similar to the implementation of the Viterbi algorithm \cite{Viterbi1989}.

The second stage is to identify the relation, which sets the edge label among the concept subgraph fragments assigned in the previous stage. The authors tackle this stage like a graph-based dependency parser problem. While the dependency parser aims to find the maximum-scoring tree over words from the sentence, the relation identifier tries to find the maximum-scoring among subgraphs that preserve concept fragments from the previous stage.

To train the two stage parser, the authors formulate the training data for concept identification and relation identification separately. In both tasks, the input must be annotated with name entities (obtained from Illinois Name Entity Tagger), part-of-speech tags and basic dependencies (obtained from Stanford Parser). The detail settings and hyper-parameters can be found in the original paper. This parser has been evaluated the first time on LDC2013E117 corpus (in 2014), archived the Smatch score of \textbf{\textit{0.58}}, and the second time on LDC2015E86 corpus (in 2016), which showed great improvement with \textbf{\textit{0.67}} Smatch score.

\subsection{Grammar-based parsing}

\begin{figure*}[h]
\centerline{\includegraphics[width=4.7in]{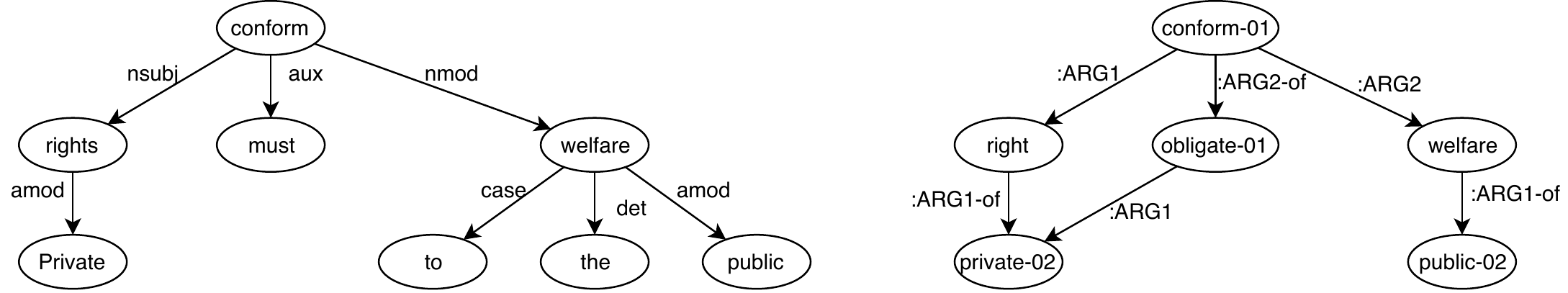}}
\caption{Dependency tree and AMR graph generated from the sentence \textit{"Private rights must conform to the public welfare"}}
\label{fig:camr}
\end{figure*}

After the success of Flanigan et al. with an alignment-based approach \cite{Flanigan_adiscriminative}, Wang et al. \cite{wang2016:SemEval} introduced a grammar-based (or transition-based) parser called \textbf{CAMR}. The authors first use a dependency parser to generate the dependency tree for the sentence, then transform the dependency tree to an AMR graph through some transition rules. This method takes advantages of the achievements in dependency parsing, with a training set much larger than the training set of AMR parsing. Damonte et al. \cite{Damonte_Incremental}. Brandt et al. \cite{Brandt2016}, Goodman et al. \cite{Goodman2016UCLSheffieldAS} and Peng et al. \cite{Peng2018AMRPW} also applied the grammar-based algorithm in their works and obtained competitive results. Figure \ref{fig:camr} shows an example of the dependency tree and the AMR graph parsed from the same sentence \textit{"Private rights must conform to the public welfare"}.

Unlike the dependency tree of a sentence, where each word corresponds to a node in the tree, in AMR graph, some words become abstract concepts or relations while other words are simply deleted because they do not contribute to the meaning. This difference causes many difficulties for aligning word tokens and the concept. In order to learn the transition from the dependency tree to AMR graph, Wang et al. \cite{wang2016:SemEval} use the algorithm from JAMR to produce the alignment. The authors also construct a span graph to represent an AMR graph that is aligned with the word tokens in a sentence. This span graph is a directed, labeled graph $G = (V, A)$, where $V = \{s_{i,j} |i, j \in (0, n)$ and $j > i$ \} is a set of nodes, and $A \subseteq V \times V$ is a set of  arcs. Each node $s_{i,j}$ in $G$ is assigned a concept label from concept set $L_V$ and is mapped with a continuous span $(w_i, ..., w_{j-1})$ in the sentence $w$. Each arc is also assigned a relation label from relation set $L_A$.

Basically, CAMR will perform three types of actions to transform the dependency tree into the AMR graph: actions performed when an edge is visited, actions performed when a node is visited, and actions used to infer abstract concepts in AMR that does not correspond to any word or word sequence in the sentence. For the details of these actions, readers can refer to the original paper \cite{Wang2015}, the Boosting version \cite{Wang2015BoostingTA} and the paper at Semeval2016 contest \cite{wang2016:SemEval}. A disadvantage of this method is that it limits the parsing ability to a single sentence, because the dependency tree can cover only the structure inside a sentence. 

Damonte et al. \cite{Damonte_Incremental} developed a transition-based model called \textbf{AMREager} that also parses the AMR graph based on transition rules, but differs from CAMR which requires the full dependency tree to be obtained and then process the tree bottom-up, this parser process the sentence left-to-right. AMREager defines a stack, a buffer and a configuration to perform the transition actions, which can be: \textit{Shift, LArc, RArc} or \textit{Reduce}. AMREager also uses the alignment obtained from JAMR aligner to map indices from the sentence to AMR graph fragments. Although the result in Smatch score is still lower than CAMR and JAMR by a small margin, AMREager obtains best results on several subtasks such as Name Ent. and Negation.

\subsection{Machine-translation-based parsing}

Recently, with the achievement of the encoder-decoder architecture in deep neural networks, several supervised learning approaches have been proposed in order to deal with AMR parsing task. They attempt to linearize the AMR in Penman notation to sequences of text, at character-level \cite{Noord2017NeuralSP} or at word-level \cite{Kontas2017} \cite{Peng2017AddressingTD}, so that the parsing task can be considered as a translation task, which transforms a sentence into an AMR-like sequence. In this paper, we choose \textbf{NeuralAMR } (word-level linearization) \cite{Kontas2017} and \textbf{Ch-AMR} (character-level linearization) \cite{Noord2017NeuralSP} to run our experiments.

\begin{table}
    \caption{AMR linearization for the sentence \textit{"Private rights must conform to the public welfare"} in NeuralAMR - the left side is the original AMR and the right side is the linearized string}
    \label{tab:neural_amr}
    \begin{center}

        \begin{tabular}{|m{5cm}|m{5cm}|}
            \hline
            (o / obligate-01 \newline \hspace*{0.3cm} :ARG1 (r / right-05 \newline \hspace*{0.6cm} :ARG1-of (p / private-02)) \newline \hspace*{0.3cm} :ARG2 (c / conform-01 \newline \hspace*{0.6cm} :ARG1 r \newline \hspace*{0.6cm} :ARG2 (w / welfare \newline \hspace*{0.9cm} :ARG1-of (p2 / public-02)))) & (obligate-01 :ARG1 (right-05 :ARG1-of (private-02)) :ARG2 (conform-01 :ARG1 (right-05) :ARG2 (welfare :ARG1-of (public-02))))\\
            \hline
        \end{tabular}
    \end{center}
\end{table}

Given an AMR graph represented in Penman notation, NeuralAMR preprocesses the graph through a series of steps: AMR linearization, anonymization, and other modifications which aim to reduce the complexity of the linearized sequences and to address sparsity from certain open class vocabulary entries, such as named entities and quantities. Representing AMR graphs in this way, NeuralAMR takes advantage of sequence-to-sequence model by using a stack bidirectional LSTM encoder to encode the input sequence and a stacked LSTM to decode from the hidden states produced by the encoder. The output string of the model is converted back to AMR format to complete the parsing process. Since this approach requires a huge amount of labeled data, NeuralAMR uses paired training procedure to bootstrap a high-quality AMR parser from millions of unlabeled Gigaword sentences. With this extra dataset, the parsing result increases significantly, from \textbf{\textit{0.55}} to \textbf{\textit{0.62}} in Smatch score. However, it is difficult for this linearization method to keep the structure of the original graph. In the example shown in Figure \ref{fig:grammar}, the distance between two nodes: \textit{"obligate-01"} and \textit{"conform-01"}, which are directly connected in the graph, becomes 5 (tokens) in the linearized string, as shown in Table \ref{tab:neural_amr}. This distance can be even larger in long sentences with complicated structure, thus causes many mistakes in the annotation. 

\begin{figure}
    \centerline{\includegraphics[width=3.5in]{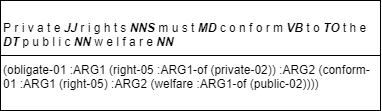}}
    \caption{Preprocessing data in Ch-AMR - The sentence is converted to sequence of characters with POS tag in uppercase following each word, the AMR graph is linearized and removed all variables}
    \label{fig:char_based}
\end{figure}

Different from Kontas et al. \cite{Kontas2017}, Noord and Bos \cite{Noord2017NeuralSP} introduce another approach in linearizing which transforms the AMR graph to the character-level. This model removes all variables from the AMRs and duplicate co-referring nodes. The input sentences are also tokenized in character-level, along with the part-of-speech tag of the original tokens to provide more linguistic information to the decoder. An example of such a preprocessed AMR is shown in Figure \ref{fig:char_based} Obviously, this preprocessing method causes losing information, since the variables cannot be put back perfectly. To tackle this limitation, the authors describe an approach to restore the co-referring nodes in the output. All wikification relations present in AMRs in the training set are also removed and restored in a post-processing step. This model archives a better result, with \textbf{0.71} Smatch score. 

\section{Dataset and evaluation}\label{dataset}
\subsection{Dataset}

The original dataset used for testing in this paper is JCivilCode-1.0, which is introduced by Lai et al. in \cite{Lai2017}. In our work, we revised JCivilCode-1.0 carefully with some modifications and extracted more 48 articles to complete the first four chapter in Part I of the Japanese Civil Code. All the AMRs are annotated by a group of annotators to ensure the neutrality of evaluation. Table \ref{tbl:jcivilcode} shows some statistics of this dataset after our revision.

As we mentioned in section \ref{intro}, one of the main difficulty in analyzing legal documents is dealing with long sentences. In our experiments, we also would like to assess the performances of the five models with different length of the sentence. Since the current legal dataset is still small, we use extra sentences extracted from the well-known LDC2017T10 dataset, which consists of nearly 40,000 sentences in news domain. We divide the test set of LDC2017T10 into four subsets LDC-20, LDC-20-30, LDC-30-40, LDC-40 with the lengths of the sentences in range 0-20, 21-30, 31-40 and greater than 40 words, respectively. We excluded the samples containing sub-sentences inside (annotated \textit{"multi-sentence"} by the annotators). This exclusion guaranteed a fair comparison among the five parsers because CAMR is unable to analyze multiple sentences at the same time.

\begin{table}
    \begin{center}
        \caption{JCivilCode-1.0 statistic}
        \begin{tabular}{l|c}
            \textbf{Number of samples} & 128 \\
            \hline
            \textbf{Average length} & 31 \\
            \hline
            \textbf{Max sentence length} & 107 \\
            \hline
            \textbf{Average number of graph nodes} & 28 \\
            \hline 
            \textbf{Max number of graph nodes} & 96 \\
            \hline
            \textbf{Vocabulary size} & 796 \\
            \hline
            \textbf{Number of tokens} & 4042 \\
        \end{tabular}
        \label{tbl:jcivilcode}
    \end{center}
\end{table}

\subsection{Evaluation}

AMR parsers are evaluated mainly by Smatch score \cite{cai2013smatch}. Given the parsed graphs and the gold graphs in the form of Penman annotations, Smatch first tries to find the best alignments between the variable names for each pair of graphs and it then computes precision, recall and F1 of the concepts and relations. In this paper, to test the performance of AMR parser on legal text, which contains sentences in complicated structures, we analyze the parsing results in a deeper measurement. Specifically, we use the test-suite introduced by Damonte et al. \cite{Damonte_Incremental}, which assesses the parsing results on various sub-score as follow:

\begin{itemize}
    \item \textit{Unlabeled}: Smatch score computed on the predicted graphs after removing all edge labels (e.g., \textit{:ARG0, :condition})
    \item \textit{No WSD}: Smatch score while ignoring Propbank senses (e.g., \textit{perform-01} vs \textit{perform-02})
    \item \textit{Name Entity}: F-score on the named entity recognition (:name roles)
    \item \textit{Wikification}: F-score on the wikification (:wiki roles)
    \item \textit{Negation}: F-score on the negation detection (:polarity roles)
    \item \textit{Concepts}: F-score on the concept identification task 
    \item \textit{Reentrancies}: Smatch computed on reentrant edges only
    \item \textit{SRL}: Smatch computed on :ARG-i roles only
\end{itemize}

In our experiment with JCivilcode-1.0, we do not include the Wikification and Name Entity criteria since there are no Wiki concepts included in this dataset, and the number of existing named entities is small.

\section{Experiments and discussion}\label{experiments}

To evaluate the performance of different parsing strategies on legal text, we conduct experiments on five models that already provided their source codes: JAMR, CAMR, AMR-Eager, NeuralAMR and Ch-AMR. While JAMR, CAMR and AMR-Eager were trained with the LDC2015E86 dataset only (the older version of LDC2017T10), NeuralAMR and Ch-AMR initialized the parser by LDC2015E86 and then used an extra corpus of 2 millions sentences extracted from a free text corpus Gigaword \cite{Gigaword} to train the complete models. We provide some statistics about LDC2015E86 and LDC2017T10 in Table \ref{tab:ldc_stats}. English sentences in these two datasets are collected from TV program transcriptions, web blogs and forums. Each sample in these datasets includes a pair of sentence and AMR graph corresponding. 

\begin{table}[]
    \centering
    \caption{LDC2015E86 and LDC2017T10 number of samples}
    \begin{tabular}{l|c|c|c|c}
         Dataset & Total & Train & Dev & Test \\
         \hline
         LDC2015E86 & 19,572 & 16,833 & 1,368 & 1,371 \\ 
         \hline
         LDC2017T10 & 39,260 & 36,521 & 1,368 & 1,371 \\ 
         \hline
         LDC-20 & - & - & - & 694 \\
         \hline
         LDC-20-30 & - & - & - & 284 \\
         \hline
         LDC-30-40 & - & - & - & 143 \\
         \hline
         LDC-40 & - & - & - & 82
    \end{tabular}
    \label{tab:ldc_stats}
\end{table}

Parsing results are summarized in Table \ref{tbl:resultldc} (LDC2017T10 long sentences experiments) and Table \ref{tbl:resultcivil} (JCivilCode1.0 experiments). Overall, the Smatch score of all the parsers on JCivilCode-1.0 is still lower than on LDC2015E86 by a large margin. It can be figured out that grammar-based and alignment-based methods showed promising results over MT-based method. JAMR and CAMR archieved the best score on LDC2017T10 long sentences and JCivilCode-1.0 dataset, respectively, while AMREager's performance was competitive on both tasks.

\begin{table}
    \begin{center}
        \caption{Smatch scores on LDC2017T10}
        \begin{tabular}{l|c|c|c|c|c}
            & \textbf{JAMR} & \textbf{CAMR} & \textbf{AMREager} & \textbf{NeuralAMR} & \textbf{Ch-AMR} \\
            \hline
            LDC-20 & \textbf{0.71} & 0.66 & 0.69 & 0.65 & 0.45 \\
            LDC-20-30 & \textbf{0.68} & 0.62 & 0.64 & 0.59 & 0.43 \\
            LDC-30-40 & \textbf{0.66} & 0.60 & 0.62 & 0.56 & 0.42 \\
            LDC-40 & \textbf{0.65} & 0.59 & 0.62 & 0.54 & 0.40 \\
            
        \end{tabular}
        \label{tbl:resultldc}
    \end{center}
\end{table}

\begin{table}
    \begin{center}
        \caption{Smatch scores and sub-scores on JCivilcode-1.0}
        \begin{tabular}{l|c|c|c|c|c}
            & \textbf{JAMR} & \textbf{CAMR} & \textbf{AMREager} & \textbf{NeuralAMR} & \textbf{Ch-AMR} \\
            \hline
            Smatch & 0.45 & \textbf{0.48} & 0.43 & 0.39 & 0.28 \\ 
            \hline
            Unlabeled & 0.50 & \textbf{0.56} & 0.53 & 0.46 & 0.37\\
            No WSD & 0.47 & \textbf{0.50} & 0.45 & 0.40 & 0.28 \\
            Negation & 0.23 & 0.16 & 0.32 & \textbf{0.35} & 0.19 \\
            Concepts & 0.59 & \textbf{0.63} & 0.62 & 0.52 & 0.35 \\
            Reentrancies & 0.32 & \textbf{0.35} & 0.31 & 0.29 & 0.22  \\
            SRL & 0.43 & \textbf{0.47} & 0.41 & 0.40 & 0.28 \\
        \end{tabular}
        \label{tbl:resultcivil}
    \end{center}
\end{table}

In LDC2017T10 experiments, JAMR remained the best parser in every range of sentence length. The gap between this method and the others even becomes larger when parsing longer sentences. Although grammar-based methods focus on constructing the structure of the graph based on its corresponding dependency tree, CAMR and AMREager are unable to provide better output than JAMR.

In legal text parsing experiments, CAMR outperforms the others on both the Smatch score and many sub-scores. Specifically, this method obtains best results in constructing graph topology (\textit{Unlabeled}), predicting the Propbank sense (No WSD and SRL) as well as identifying concepts in AMR graphs (Concepts score). When parsing graphs containing cycles, CAMR also performs better, as shown in Reentrancies scores.

\begin{table}[htbp]
\vspace{0.5cm}
    \centering
    \caption{
    Common errors types:
    \textbf{\textcolor{mygreen}{Incorrect concept}} - \textbf{\textcolor{myblue}{Incorrect relation}} - \textbf{\textcolor{myyellow}{Missing concept}} - 
    \textbf{\textcolor{myred}{Missing attribute}} 
    }
    \begin{tabular}{|p{2cm}|p{5.3cm}|p{4.8cm}|}
        \hline
        Example & \textit{Private rights must conform to the public welfare} (1) & \textit{No abuse of rights is permitted} (2) \\
        \hline
        Gold annotation & (o / obligate-01 \newline \hspace*{0.3cm} :ARG1 (r / right-05 \newline \hspace*{0.6cm} :ARG1-of (p / private-02)) \newline \hspace*{0.3cm} :ARG2 (c / conform-01 \newline \hspace*{0.6cm} :ARG1 r \newline \hspace*{0.6cm} :ARG2 (w / welfare \newline \hspace*{0.9cm} :ARG1-of (p2 / public-02)))) & (p / permit-01 \newline \hspace*{0.3cm} :polarity - \newline \hspace*{0.3cm} :ARG1 (a / abuse-01 \newline \hspace*{0.6cm} :ARG1 (r / right-05))) \\
        \hline
        JAMR & (c / conform-01 \newline \hspace*{0.3cm} :ARG1 (r / \textbf{\textcolor{mygreen}{right}} \newline \hspace*{0.6cm} :ARG1-of (p2 / \textbf{\textcolor{mygreen}{private-03}}))  \newline \hspace*{0.3cm} :ARG2 (w / welfare  \newline \hspace*{0.6cm} \textbf{\textcolor{myblue}{:domain-of}} (p / \textbf{\textcolor{mygreen}{public}}))) \newline \newline (missing concept \textbf{\textcolor{myyellow}{“obligate-01”}}) & (p / permit-01 \newline \hspace*{0.3cm} :ARG1 (a / abuse-01  \newline \hspace*{0.6cm} :ARG1 (r / \textbf{\textcolor{mygreen}{right}}))  \newline \hspace*{0.3cm} :polarity -) \\
        \hline
        CAMR & (x2 / right-05 \newline \hspace*{0.3cm} :ARG1-of (x1 / \textbf{\textcolor{mygreen}{private-03}})  \newline \hspace*{0.3cm} :ARG1-of (x4 / conform-01 \newline \hspace*{0.6cm} :ARG2 (x8 / welfare \newline \hspace*{0.9cm} \textbf{\textcolor{myblue}{:mod}} (x7 / \textbf{\textcolor{mygreen}{public}})))) \newline \newline (missing concept \textbf{\textcolor{myyellow}{“obligate-01”}}) & (x6 / permit-01 \newline \hspace*{0.3cm} :ARG1 (x2 / abuse-01  \newline \hspace*{0.6cm} :ARG1 (x4 / \textbf{\textcolor{mygreen}{right}}))) \newline \newline (missing attribute \textbf{\textcolor{myred}{“:polarity -”}}) \\
        \hline
        AMR-Eager & (v3 / conform-01 \newline \hspace*{0.3cm} :ARG1 (v2 / \textbf{\textcolor{mygreen}{right}} \newline \hspace*{0.6cm} :ARG1-of (v1 / \textbf{\textcolor{mygreen}{private-03}})) \newline \hspace*{0.3cm} :ARG2 (v5 / welfare \newline \hspace*{0.6cm} \textbf{\textcolor{myblue}{:mod}} (v4 / \textbf{\textcolor{mygreen}{public}}))) \newline \newline (missing concept \textbf{\textcolor{myyellow}{“obligate-01”}}) & (v3 / permit-01 \newline \hspace*{0.3cm} :ARG1 (v1 / abuse-01  \newline \hspace*{0.6cm} polarity - \newline \hspace*{0.3cm} :ARG1 (v2 / \textbf{\textcolor{mygreen}{right}}))) \\
        \hline
        Neural-AMR & (o / obligate-01 \newline \hspace*{0.3cm} :arg2 (r / \textbf{\textcolor{mygreen}{rule-out-02}} \newline \hspace*{0.6cm} :arg0 (r2 / \textbf{\textcolor{mygreen}{right}} \newline \hspace*{0.9cm} :arg1-of (p / \textbf{\textcolor{mygreen}{private-03}})) \newline \hspace*{0.6cm} :arg1 (w / welfare \newline \hspace*{0.9cm} :mod (p2 / public)))) & (p / permit-01 \newline \hspace*{0.3cm} :polarity -  \newline \hspace*{0.3cm} :arg1 (a / \textbf{\textcolor{mygreen}{abuse-02}} \newline \hspace*{0.6cm} :arg1 (r / \textbf{\textcolor{mygreen}{right}})))  \\
        \hline
        Ch-AMR & (vv1conform-01 / conform-01 \newline \hspace*{0.3cm} :ARG1 (vv1person / \textbf{\textcolor{mygreen}{person}} \newline \hspace*{0.6cm} :ARG1-of (vv1private-03 / \textbf{\textcolor{mygreen}{private-03}})) \newline \hspace*{0.3cm} :ARG2 (vv1welfare / welfare \newline \hspace*{0.6cm} :ARG1-of \textbf{\textcolor{mygreen}{vv1}})) \newline \newline (missing concept \textbf{\textcolor{myyellow}{"right-05", "public-02", "obligate-01"}} ) & (vv3permit-01 / permit-01 \newline \hspace*{0.3cm} :ARG1 (vv3no-abuse / \textbf{\textcolor{mygreen}{no-abuse}})) \newline \newline (missing concept \textbf{\textcolor{myyellow}{"right-05"}}) \\
        \hline
      
    \end{tabular}
    \label{tab:output}
\end{table}

\begin{figure}
\centerline{\includegraphics[width=3.3in]{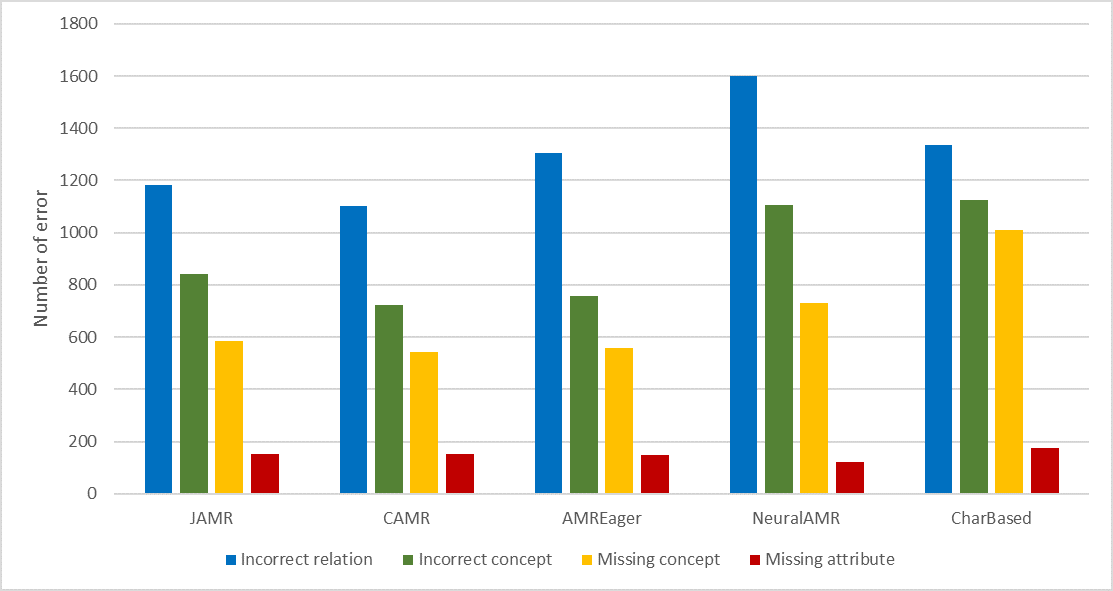}}
\caption{Common error types}
\label{fig:error}
\end{figure}

We analyze some common errors in parsing outputs of legal sentences, with the statistic represented in Figure \ref{fig:error} and the examples provided in Table \ref{tab:output}. One of the most common error in alignment-based and grammar-based performances is missing concept and relation related to \textit{modal verbs}. In legal documents, modal verbs (e.g. "may", "can", "must") play a crucial role in a sentence and decide whether an action is permitted or not. This differs from other domains, where these words do not often contribute a lot to the sentence meaning. As shown in example 1 in Table \ref{tab:output}, only NeuralAMR is capable of identifying the concept \textit{"obligate-01"} while other models totally ignore it. 

Another challenge in parsing legal text is the logical complexity. In this aspect, all the parsers still show limitation when parsing negative clause. This is not too surprising as many negations are encoded with morphology (e.g., such as \textit{“un-”} prefix in \textit{"unless"} or \textit{"unable"}) and cause difficulties for detection. We show an example in Table \ref{tab:output} the output from all the parsers for a sentence: \textit{"No abuse of rights is permitted"}. NeuralAMR and JAMR succeeded in converting negation to \textit{:polarity -}, AMREager didn't put this edge to the exact position, but in this case, it doesn't change the meaning of the sentence. CAMR even performs worse as it skips this important information.

\section{Conclusions}

We conducted experiments of AMR parsing on the legal dataset JCivilCode-1.0 and news domain dataset LDC2017T10 with different ranges of sentence length to observe the abilities of five different models. The parsing outputs were evaluated by Smatch metric in several aspects including overall F-score and sub-score on specific tasks. Experimental results showed the domain adaptation of five models for the legal domain and the performance decreased of approximately 0.2 on the Smatch score. This result shows difficulties in applying AMR parsing for analyzing legal documents. 

Currently, our legal dataset JCivilcode is still too small comparing to LDC2017T10. In order to improve the domain adaptation ability for current approaches as well as to obtain a fully evaluation, the legal dataset has to be enlarged. This work requires a lot of efforts from experts in both linguistic and legal domain.

\section*{Acknowledgment}
This work was supported by JST CREST Grant Number JPMJCR1513, Japan.

%
%
%
%
\bibliographystyle{splncs04}
\bibliography{references/jurisin2018.bib} 
\end{document}